\documentclass[preprint,12pt]{elsarticle}

\usepackage{url}

\usepackage{amssymb}

\journal{Information Sciences}

\begin{document}

\begin{frontmatter}



\title{Why `GSA: A Gravitational Search Algorithm'\\ Is Not Genuinely Based on the
Law of Gravity}


\author{Melvin Gauci, Tony J. Dodd, and Roderich Gro{\ss}}

\address{Department of Automatic Control and Systems Engineering\\ 
The University of Sheffield, Mappin Street, S1 3JD Sheffield, UK}



\end{frontmatter}

In~\cite{rashedi09:_gsa}, Rashedi, Nezamabadi-pour and Saryazdi
propose an interesting optimization algorithm called `GSA: A Gravitational Search
Algorithm'. The paper has received numerous citations since its
publication in 2009.

GSA simulates a set of agents in an $N$--dimensional search space, which
behave as point masses. The position of agent $i$, denoted by
$\vec{x}_i$, represents a candidate solution for the optimization
problem at hand. The mass of agent $i$, denoted by $m_i$, is
correlated with the quality of the candidate solution it represents:
the higher the quality, the higher the mass. The agents are said to
``interact with each other based on the Newtonian gravity and the laws
of motion''~\cite{rashedi09:_gsa}. If the law of gravity were to be
strictly followed, the \textit{magnitude} of the force exerted on
agent $i$ by agent $j$ (and vice versa) should be ``proportional to
the mass of each and [vary] inversely as the square of the distance
between them''~\cite{Feynman2005}.  Formally,
\begin{equation}
  ||\vec{f}_{ij}{||}=G\frac{m_i  m_j}{R_{ij}^2}\label{eq:inverse_square_law},
\end{equation}
where $G$ is a proportionality constant analogous to the universal
gravitational constant and $R_{ij}=||\vec{x}_j-\vec{x}_i{||}$ is the Euclidean distance
between agents $i$ and $j$.

The authors claim to have implemented a law of gravity where 
the magnitude of the force
between two agents is inversely proportional to the distance $R$
between them. They use ``$R$ instead of $R^2$, because according to
[their] experiment results, $R$ provides better results than $R^{2}$
in all experimental cases''~\cite{rashedi09:_gsa}.

In GSA, according to Equation 7 in~\cite{rashedi09:_gsa}, the force exerted on
agent $i$ by agent $j$ along dimension $d\in\{1, 2, \dots, N\}$ of the
search space is calculated as follows:
\begin{equation}
f_{ij}^{d}=G\frac{m_{i}m_{j}}{R_{ij}+\epsilon}\left(x_{j}^{d}-x_{i}^{d}\right)\label{eq:wrong_component},
\end{equation}
where $\epsilon$ is a small positive constant used to avoid division by zero.


However, in order to obtain the force vector $\vec{f}_{ij}$, its magnitude, $||\vec{f}_{ij}||$, should be multiplied
by the \textit{unit} vector pointing from agent $i$ to agent $j$.
Equation \ref{eq:wrong_component} is not correct in that
$\left(x_{j}^{d}-x_{i}^{d}\right)$ is not normalized. The denominator
of Equation \ref{eq:wrong_component} should read $R_{ij}^{3}$ in order
to implement an inverse-square law, or $R_{ij}^{2}$ in order to
implement an inverse-linear law (as the authors intended). Using
$R_{ij}$ effectively makes the magnitude of the force $\vec{f}_{ij}$ independent of
the distance between agents $i$ and $j$. Formally, assuming
$\epsilon=0$,
\begin{equation}
||\vec{f}_{ij}||= Gm_im_j.
\label{eq:wrong2}
\end{equation}

We do not in any way intend to undermine GSA's capabilities as an
optimization algorithm; rather, our observation is that it is not 
strictly motivated by how gravity works (compare
Equations~\ref{eq:inverse_square_law} and~\ref{eq:wrong2}). Mass and distance are 
both integral parts of the law of gravity;
in contrast, the force model used in GSA is independent of the distance between the agents and
is based solely on their masses (qualities of candidate solutions).

\bibliographystyle{model1-num-names}
\bibliography{paper}







\end{document}